\pdfoutput=1

\documentclass[11pt]{article}

\usepackage{acl}

\usepackage{times}
\usepackage{latexsym}
\usepackage{todonotes}
\usepackage{amsmath}
\usepackage[T1]{fontenc}

\usepackage[utf8]{inputenc}
\usepackage{float}
\restylefloat{table}
\usepackage{microtype}
\usepackage{graphicx}
\usepackage{algorithm}
\usepackage{algorithmic}
\usepackage{amssymb}
\usepackage{multirow}
\usepackage{enumitem}
\usepackage{dirtytalk}
\title{PESE: Event Structure Extraction using Pointer Network based Encoder-Decoder Architecture}

\author{Alapan Kuila \\
  IIT Kharagpur, India \\
  \texttt{alapan.cse@iitkgp.ac.in} \\\And
  Sudeshna Sarkar \\
  IIT Kharagpur, India\\
  \texttt{sudeshna@cse.iitkgp.ac.in} \\}

\begin{document}
\maketitle
\begin{abstract}
The task of event extraction (EE) aims to find the events and event-related argument information from the text and represent them in a structured format. Most previous works try to solve the problem by separately identifying multiple substructures and aggregating them to get the complete event structure. The problem with the methods is that it fails to identify all the interdependencies among the event participants (event-triggers, arguments, and roles). In this paper, we represent each event record in a unique tuple format that contains trigger phrase, trigger type, argument phrase, and corresponding role information. Our proposed pointer network-based encoder-decoder model generates an event tuple in each time step by exploiting the interactions among event participants and presenting a truly end-to-end solution to the EE task. We evaluate our model on the ACE2005 dataset, and experimental results demonstrate the effectiveness of our model by achieving competitive performance compared to the state-of-the-art methods.
\end{abstract}

\section{Introduction}

Event extraction (EE) from text documents is one of the crucial tasks in natural language processing and understanding. Event extraction deals with the identification of event-frames from natural language text. These event-frames have a complex structure with information regarding event-trigger, event type, event-specific arguments, and event-argument roles. For example, 
\begin{quote}
In Baghdad, a cameraman \textbf{died} when an American tank \textbf{fired} on the Palestine hotel.
\end{quote}
In this sentence \textbf{died} and \textbf{fired} are the event triggers for the event types \textit{Die} and \textit{Attack} respectively. The sentence contains entities phrases: \textit{Baghdad, a cameraman, an American tank} and \textit{Palestine hotel}. Some of these entities play a specific role in these mentioned events and termed as event arguments. For event type \textit{Die}, (argument; role) pairs are: (Baghdad; Place), (A cameraman; victim), (American tank; instrument). Whereas, for \textit{Attack} event, (argument; role) pairs are:
(Baghdad; Place), (A cameraman; target), (American tank; instrument), and (Palestine Hotel; Target). Apparently, a sentence may contain multiple events; an entity may be shared by multiple event frames; moreover, a specific argument may play different roles in different event frames. Therefore an ideal event extraction system will identify all the trigger words, classify the correct event types, extract all the event-specific arguments and correctly predict the event-argument roles. Each of these subtasks is equally important and challenging. 

Most existing works decompose the EE task into these predefined subtasks and later aggregate those outputs to get the complete event frames. Some of these models follow a pipelined approach where triggers and corresponding arguments are identified in separate stages. In contrast, others rely on joint modeling that predicts triggers and relevant arguments simultaneously. However, the pipeline approaches have to deal with error propagation problems, and the joint models have to exploit the information sharing and inter-dependency among the event triggers, arguments, and corresponding roles. The interaction among the event participants are of the following types: 1) inter-event interaction: usually event types in one sentence are interdependent of one another~\cite{Chen2018CollectiveED} 2) intra-event argument interaction: arguments of a specific event-mention have some relationship among themselves~\cite{Sha2016RBPBRP}~\cite{Sha2018JointlyEE}~\cite{hong-etal-2011-using} 3) inter-event argument interaction: target entities or arguments shared by two different event mention present in a sentence generally have some inter-dependencies~\cite{hong-etal-2011-using}~\cite{Nguyen2016JointEE} 4) event type-role interaction: Each event frame has a distinct set of argument roles based on its schema definition; hence event type and argument roles have an assiduous relationship.~\cite{DBLP:conf/acl/Xi0ZWJW20} 5) argument-role interaction: the event-argument role is dependent on the entity types of the candidate arguments~\cite{DBLP:conf/acl/Xi0ZWJW20} as well. Significant efforts have been devoted to exploiting these interactions but despite their 
promising results, most of these existing systems failed to capture all these inter-dependencies~\cite{DBLP:conf/acl/Xi0ZWJW20}~\cite{Nguyen_Nguyen_2019}.


In order to exploit the interactions among the event participants mentioned above, we propose a neural network-based sequence to structure learning model that can generate sentence-level event frames from the input sentences.  Each event frame holds a (trigger, argument) phrase pair along with corresponding trigger type(event type) and role-label information. 
Inspired from the models used for joint entity-relation extraction ~\cite{nayak2019ptrnetdecoding}~\cite{chen-etal-2021-jointly}, aspect sentiment triplet extraction ~\cite{mukherjee-etal-2021-paste} and semantic role labeling ~\cite{Fei2021EncoderDecoderBU}, we design a \textbf{P}ointer network-based \textbf{E}vent \textbf{S}tructure \textbf{E}xtraction (PESE) framework~\footnote{codes are available at \url{https://github.com/alapanju/PESE.git}} that utilizes the event-argument-role interdependencies to extract the event frames from text. The encoder encodes the input sentence, whereas the decoder identifies an event frame in each time step based on the input sentence encoding and the event frames generated in the previous time steps. The innovation lies in the effectiveness of this type of modeling: 1) instead of decomposing the whole task into separate subtasks, our model can detect the trigger, argument, and role labels together 2)The system is capable of extracting multiple events present in a single sentence by generating each event-tuple in consecutive time steps, 3) the model is also able to extract multiple event-tuples with common trigger or argument phrases and 4)experimental results show that the model can identify the overlapping argument phrases present in the sentence as well. In summary, the contributions of this paper are:

(1) We propose a new representation schema for event frames where each frame contains information regarding an (event, argument) phrase pair. 

(2) We present a sentence-level end-to-end event extraction model which exploits the event-argument-role inter-relatedness and tries to find the trigger, argument spans, and corresponding labels within a sentence. The proposed EE system takes a sentence as input and generates all the unique event frames present in that sentence as output.

(3) We have applied our proposed method to the ACE2005 dataset\footnote{\url{https://catalog.ldc.upenn.edu/LDC2006T06}} and the experimental results show that our approach outperforms several state-of-the-art baselines models.

\section{Event Frame Representation}
\label{etr}
Given a sentence, our proposed end-to-end EE model extracts all the event-frames present in that sentence. These event frames are the structured representation of the event-specific information: (1) Event trigger phrase, (2) Event type, (3) Argument phrase, (4) Role label. Inside the sentences, each trigger and argument phrase appears as a continuous sequence of words; hence, an effective way to represent these phrases is by their corresponding start and end locations. Therefore in this paper, we represent each event-frame using a 6-tuple structure that stores all the records, as mentioned earlier. The 6-tuple contains: 1) start index of trigger phrase, 2) end index of trigger phrase, 3) event type, 4) start index of argument phrase, 5) end index of argument phrase 6) trigger-argument role label. The start and end index of the trigger phrase(1-2) denotes the event-trigger span, whereas the start and end index of argument phrase(4-5) represent the event-argument span and the other two records(3 and 6) are two labels: event type and role type. Table~\ref{table:frame} represents sample sentences and corresponding event frames present in those sentences with their 6-tuple representations. However, there are instances when an event-trigger is present in a sentence without any argument phrase. In order to generalize the event-tuple representation, we concatenate two extra tokens: \verb|[unused1]| and \verb|[unused2]| in front of each sentence with position $1^{st}$ and $2^{nd}$ respectively~\ref{table:frame}. In the absence of an actual argument phrase in the sentence, the \verb|[unused2]| token is used as the dummy argument, and the corresponding start and end index of the argument phrase in the event-tuple are represented by $1$, and the role-type is represented by \say{NA} (see Table~\ref{table:frame}). The token \verb|[unused1]| is used to indicate the absence of any valid event-trigger word in the sentence.

\begin{table*}
\center\fontsize{9}{11}\selectfont

\setlength\tabcolsep{2pt}
\begin{tabular}{|l|l|}
\hline
Input Sentence & \begin{tabular}[c]{@{}l@{}}{[}unused1{]} {[}unused2{]} Orders went out today to deploy 17,000 U.S. Army soldiers\\  in the Persian Gulf region .\end{tabular}                           \\ \hline
Output Tuple   & 7 7 Movement:Transport 8 11 Artifact , 7 7 Movement:Transport 13 16 Destination                                                                                                 \\ \hline \hline
Input Sentence & \begin{tabular}[c]{@{}l@{}}{[}unused1{]} {[}unused2{]} The more they learn about this invasion , the more they learn\\  about this occupation , the less they support it .\end{tabular} \\ \hline
Output Tuple   & 8 8 Conflict:Attack 1 1 NA                                                              \\ \hline
\end{tabular}
\caption{Event tuple representation for Encoder decoder model}
\label{table:frame}
\end{table*}

\subsection{Problem Formulation}
To formally define the EE task, first we consider two predefined set E and R where 
$E \in \{E_1, E_2, E_3,\ldots, E_p\}$ is the set of event types, 
and $R \in \{R_1, R_2, R_3,\ldots, R_r\}$ is the set of role labels. Here $p$ and $r$ are number of event types and role types respectively. Now, given a sentence $S = [w_1, w_2, w_3,..., w_n]$ where $n$ is the sentence length and $w_i$ is the $i$th token, our objective is to extract a set of event-tuples $ET = \left\{et_i\right\}_{i=1}^{|ET|}$ where $et_i = [s_i^{tr}, e_i^{tr}, E_i, s_i^{ar}, e_i^{ar}, R_i]$ and $|ET|$ indicates number of event frames present in sentence $S$. In the $i$th event-tuple ($et_i$) representation, $s_i^{tr}$ and $e_i^{tr}$ respectively represent the start and end index of trigger phrase span, $E_i$ indicates the event type of the candidate trigger from set $E$, $s_i^{ar}$ and $e_i^{ar}$ respectively denote the start and end index of argument phrase span and $R_i$ indicates role-label of the (trigger, argument) pair from set $R$.

\section{Our Proposed EE Framework}

\begin{figure}
  \includegraphics[width=0.49\textwidth]{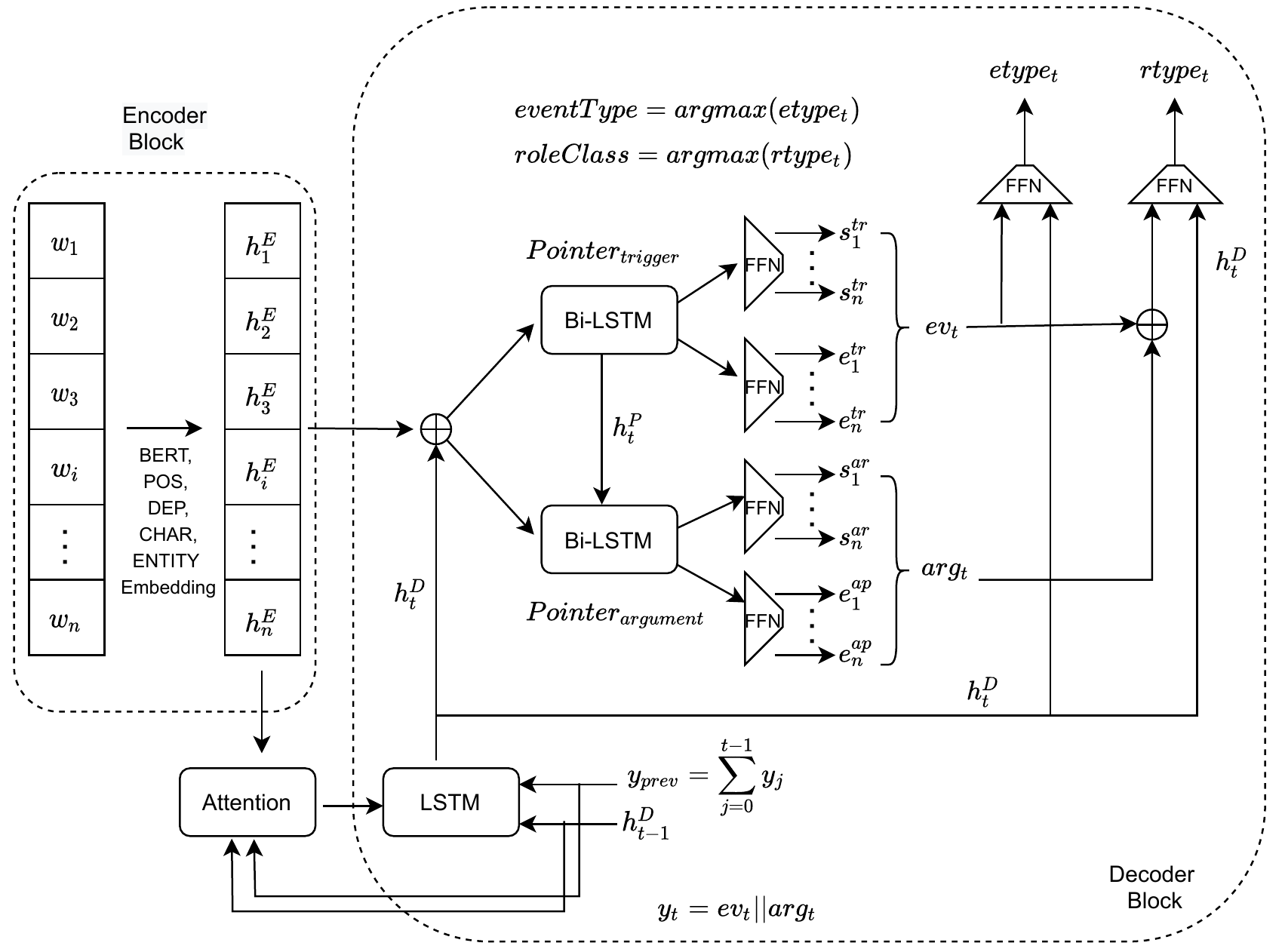}
  \caption{Pointer network based encoder decoder model architecture}
  \label{fig:arch}
\end{figure}

We employ a encoder-decoder architecture for the end-to-end EE task. The overview of the model architecture is depicted in Figure~\ref{fig:arch}. The input to our model is a sentence (i.e. a sequence of tokens) and as output, we get a list of event tuples present in that sentence. We use pre-trained BERT \cite{Devlin2019BERTPO} at the encoder and LSTM \cite{10.1162/neco.1997.9.8.1735}-based network at the decoder in our model. 


\subsection{Sentence Encoding} 

We use pre-trained BERT model as the sentence encoder to obtain the contextual representation of the tokens. However, part-of-speech (POS) tag information is a crucial feature as most trigger phrases are nouns, verbs or adjectives. Besides, the dependency tree feature (DEP) is another informative clue in sentence-level tasks~\cite{Sha2018JointlyEE}. We also use the entity type information (ENT) information (BIO tags) as feature. We combine the POS, DEP, ENT, and character-level features with the BERT embeddings to represent each token in the input sentence. So along with pre-trained BERT embedding we use four other embeddings: 1) POS embeddings $E_{pos} \in \mathbb{R}^{|POS|\times d_{pos}}$ 2) DEP embeddings 
$E_{dep} \in \mathbb{R}^{|DEP|\times d_{dep}}$ 3) Entity type embeddings $E_{ent} \in \mathbb{R}^{|ENT|\times d_{ent}}$ and 4) character-level embeddings 
$E_{char} \in \mathbb{R}^{|V_c|\times d_{char}}$. Here, $|POS|$, $|DEP|$, $|ENT|$ and $|V_c|$ indicates respectively the count of unique pos tags, dependency relation tags, entity tags and unique character alphabets. Whereas, $d_{pos}$, $d_{dep}$, $d_{ent}$ and $d_{char}$ represents the corresponding dimensions of pos, dependency, entity and character features respectively. Similar to ~\cite{chiu-nichols-2016-named} we apply convolution neural network with max-pooling to obtain the character-level feature vector of dimension $d_c$ for each token in the sentence $S$. All these feature representations are concatenated to get the aggregated vector representation $h_i^E$ for each token $w_i$ present in the sentence $S$. More specifically, $h_i^E \in \mathbb{R}^{d_{h}}$ where $d_{h} = d_{BERT}+d_{pos}+d_{dep}+d_{ent}+d_{c}$.


\subsection{Extraction of Event Frames} 

Our proposed decoder generates a sequence of event tuples. The decoder comprises sequence-generator LSTM, two pointer networks, and two classification networks. The event frame sequence is generated by the sequence-generator LSTM. The trigger and argument spans of the events are identified by the pointer networks. The classification networks determine the type of event and the trigger-argument role label. Each of these modules is described in greater detail below.


\paragraph{Sequence Generating Network}

We use an LSTM cell to generate the sequence of the events frame. In each time step $t$, this LSTM takes attention weighted sentence embedding ($e_t$) and aggregation of all the previously generated tuple embeddings ($eTup_{prev}$) as input and generates an intermediate hidden representation $h_t^D$($\in \mathbb{R}^{d_h}$). To obtain the sentence embedding $e_t \in \mathbb{R}^{d_h}$, we use an attention mechanism depicted in ~\cite{Bahdanau2015NeuralMT} where we use both $h_{t-1}^D$ and $eTup_{prev}$ as the query. The hidden state of the decoder-LSTM is represented as:
\begin{equation*}
    h_t^D = LSTM(e_t \oplus eTup_{prev}, h_{t-1}^D)
\end{equation*}
While generating the present tuple, we consider the previously generated tuple representations with the aim to capture the event-participant's inter-dependencies and to avoid generation of duplicate tuples.
The sentence embedding vector $e_t$ is generated by applying attention method depicted later. The aggregated representation of all the event tuples generated before current time step $eTup_{prev} = \sum_{k=0}^{t-1}eTup_k$ where $eTup_0$ is a zero tensor. The event tuple generated at time step $t$ is represented by $eTup_t = tr_t \oplus ar_t$, where $tr_t$ and $ar_t$ are the vector representations of the trigger and entity phrases respectively that are acquired from the pointer networks (depicted later) at time step $t$. Here, $\oplus$ represents concatenation operation. While generating each event tuple, we consider these previously generated event tuples to capture the event-event inter-dependencies.

\paragraph{Pointer Network for Trigger/Argument Span Detection}
The pointer networks are used to identify the trigger and argument phrase-span in the source sentence. Each pointer network contains a Bi-LSTM network followed by two feed-forward neural networks. Our architecture contains two such pointer networks to identify the start and end index of the trigger and argument phrases respectively. In each time step $t$, we first concatenate the intermediate vector $h^D_t$ (obtained from previous LSTM layer) with the hidden vectors $h^E_i$ (obtained from the encoder) and feed them to the Bi-LSTM layer with hidden dimension $d_p$ of the first pointer network.
 The Bi-LSTM network produces a hidden vector $h^{pt}_i \in \mathbb{R}^{2d_p}$ for each token in the input sentence. These hidden representations are simultaneously passed to two feed-forward networks with a softmax layer to get two normalized scalar values ($\hat{s}^{tr}_i$ and $\hat{e}^{tr}_i$) between 0 and 1 for each token in the sentence. These two values represent the probabilities of the corresponding token to be the start and end index of the trigger phrase of the current event tuple.
\begin{equation*}
    s^{tr}_i = W^1_s*h^{pt}_i + b^1_s,\;\; 
    \hat{s}^{tr} = softmax(s^{tr})
\end{equation*}
\begin{equation*}
    e^{tr}_i = W^1_e*h^{pt}_i + b^1_e,\;\;
    \hat{e}^{tr} = softmax(e^{tr}) 
\end{equation*}

Here, $W^1_s \in \mathbb{R}^{2d_p \times 1}$, $W^1_e \in \mathbb{R}^{2d_p \times 1}$, $b^1_s$ and $b^1_e$ represents the weight and bias parameters of the first pointer network.

The second pointer network that extracts the argument phrase of the tuple also contains a similar Bi-LSTM with two feed-forward networks. At each time step, we concatenate the hidden vector $h^{pt}_i$ from the previous Bi-LSTM network with $h^D_t$ and $h^E_i$ and pass them to the second pointer network, which follows similar equations as the first pointer network to obtain $\hat{s}^{ar}_i$ and $\hat{e}^{ar}_i$. These two scalars represent the normalized probability scores of the $i$th source token to be the start and end index of the argument phrase. We consider feeding the trigger pointer network's output vector to the argument pointer network's input to exploit the trigger-argument inter-dependencies. However, the normalized probabilities $\hat{s}^{tr}_i$, $\hat{e}^{tr}_i$, $\hat{s}^{ar}_i$ and $\hat{e}^{ar}_i$ collected from the two pointer networks are used to get the vector representations of the trigger and argument phrase, $ev_t$ and $arr_t$:

\begin{equation*}
    ev_t = \sum_{i=1}^n{\hat{s}_i^{tr}*h_i^{pt}} \oplus \sum_{i=1}^n{\hat{e}_i^{tr}*h_i^{pt}}
\end{equation*}
\begin{equation*}
    arg_t = \sum_{i=1}^n{\hat{s}_i^{ar}*h_i^{pa}}\oplus \sum_{i=1}^n{\hat{e}_i^{ar}*h_i^{pr}}
\end{equation*}


\paragraph{Feed-Forward Layer for Classification}
We require two feed-forward neural network-based classification layers to identify the event type, argument type and role label in each event tuple. First, we concatenate the vector representation of trigger phrase $ev_t$ with $h_t^D$ and feed the aggregated vector to the first classification layer followed by a softmax layer to find the correct event type of the detected trigger phrase. 
\begin{equation*}
    eType_t = softmax(W_{tr}(ev_t \oplus h_t^D)+b_{tr})
\end{equation*}
\begin{equation*}
    \overline{eType}_t =argmax(\hat{eType}_t)
\end{equation*}


Finally, the concatenation of $ev_t$, $arg_t$ and $h_t^D$ are feed to the second feed-forward network followed by a softmax layer to predict the correct argument-role label while exploiting the event-role and argument-role inter-dependencies.

\begin{equation*}
    rType_t = softmax(W_{r}(ev_t \oplus arg_t \oplus h_t^D)+b_{r})
\end{equation*}
\begin{equation*}
    \overline{rType}_t =argmax(\hat{rType}_t)
\end{equation*}

\subsection{Training Procedure}
To train our model, we minimize the sum of negative log-likelihood loss for identifying the four position-indexes of the corresponding trigger and argument spans and two classification tasks: 1)event type classification and 2) role classification.

\begin{equation*}
\begin{split}
    Loss = - \frac{1}{B\times ET} \sum_{b=1}^B\sum_{et=1}^{ET}[\log(s_{b,et}^{tr}, e_{b,et}^{tr})+ \\log(s_{b,et}^{ar}, e_{b,et}^{ar})+log(eType_{b,et})\\
    +log(rType_{b,et})]
\end{split}
\end{equation*}

Here, $B$ is the batch size and $ET$ represents maximum number of event-tuples present in a sentence, $b$ indicates $b$th training instance and $et$ referes to the $et$th time step. Besides, $s_{*,*}^*$, $e_{*,*}^{*}$, $eType_{*,*}$ and $rType_{*,*}$ are respectively represents the normalized softmax score of the true start and end index location of the trigger and entity phrases and their corresponding event type and role label.

\subsection{Inference of Trigger/Argument span}
At each time step $t$, the pointer decoder network gives us four normalized scalar scores: $\hat{s}^{tr}_i$, $\hat{e}^{tr}_i$, $\hat{s}^{ar}_i$ and $\hat{e}^{ar}_i$ denoting the probability of $i$th token to be the start and end index of trigger and argument span respectively. Similarly, for each token in the source sentence $S$ (of length $n$) we get a set of four probability scores based on which the valid trigger and argument span will be extracted. We identify the start and end position of the trigger and argument phrase such that the aggregated probability score is maximized with the constraint that within an event-tuple the trigger phrase and argument phrase does not have any overlapping tokens and $1\leqslant b\leqslant e \leqslant n$ where $b$ and $e$ are the start and end position of the corresponding phrase and n is the length of the sentence. First, we choose the beginning($b$) and end($e$) position index of the trigger phrase such that: $\hat{s}^{tr}_{b} \times \hat{e}^{tr}_{e}$ is maximum. Similarly, we select the argument phrase's beginning and end position index so that the extracted argument phrase does not overlap with the event phrase span. Hence, we get four position indexes with their corresponding probability scores. We repeat the whole process, but by interchanging the sequence, i.e., first, the argument span is identified, followed by the trigger phrase span. Thus we will obtain another set of four position indexes with corresponding probability scores. To identify the valid trigger and argument phrase span, we select that index set that gives the higher product of probability scores.


\section{Experiments}
\subsection{Dataset}
The ACE2005 corpus used in this paper contains a total of $599$ documents. We use the same data split as the previous works~\cite{li-etal-2013-joint}. The training data contains $529$ documents ($14669$ sentences), validation data includes $30$ documents ($873$ sentences) and the test data consists of $40$ articles ($711$ sentences). The corpus contains $33$ event subtypes, $13$ types of arguments, and $36$ unique role labels.
Here we are dealing with a sentence-level event extraction task i.e., our proposed system finds event-frames based on the information present in the sentences. There are three types of  sentences that exist in the dataset: 
\begin{itemize}[noitemsep]
    \item Single trigger with no argument: Sentence contains only one event trigger and no argument information.
    \item Single event and related arguments: Sentence contains only one event trigger and related argument information.
    \item Multiple event and related arguments: Sentence contains more than one event trigger (of the same or different event types) with corresponding argument phrases. Each of the arguments plays the same or different roles for the mentioned triggers.
    \item No information: These sentences do not contain any event trigger corresponding to predefined event types. 
\end{itemize}
For preprocessing, tokenization, pos-tagging, and generating dependency parse trees, we use spaCy library\footnote{\url{https://github.com/explosion/spaCy}}.
 The model variant that achieves the best performance ($F_1$ score) in the validation dataset, is considered for final evaluation on the test dataset.

\subsection{Parameter Settings}
In the encoder section of our model we adopt cased version of pre-trained BERT-base model~\cite{Devlin2019BERTPO}. Similar to Bert base model, the token embedding length($d_{BERT}$) is $768$. We set the dimension of the POS embedding dimension ($d_{pos}$)$= 50$, DEP feature embedding dimension ($d_{dep}$)$= 50$, Entity feature embedding dimension ($d_{ent}$)$= 50$, character embedding dimension ($d_{char}$ $= 50$) and character-level token embedding dimension ($d_c$ $= 50$). The CNN layer that is used to extract character-level token embedding has filter size = $3$ and consider tokens with maximum length =$10$. We also set the hidden dimension of the decoder-LSTM ($d_{h}$)$= 968$ and hidden dimension of the Bi-LSTM in pointer networks ($d_{p}$)$= 968$. The model is trained for $40$ epochs with batch size $32$ and we use Adam optimizer with learning rate $0.001$ and weight decay $10^{-5}$ for parameter optimization. We set dropout probability to $0.50$ to avoid overfitting. In our experiments we use P100-PCIE 16GB GPU and total number of parameters used is $\approx 220M$. The model variant with the highest $F_1$ score on development dataset is selected for evaluation on the test data. We adopt the same correctness metrics as defined by the previous works~\cite{li-etal-2013-joint} ~\cite{Chen2015EventEV} to evaluate the predicted results.


\subsection{Baselines}
 In order to evaluate our proposed model we compare our performance with some of the SOTA models that we consider as our baseline models:
\begin{enumerate}[noitemsep]
    \item JointBeam ~\cite{li-etal-2013-joint}: Extract events based on structure prediction by manually designed features.
     \item DMCNN ~\cite{Chen2015EventEV}: Extract triggers and arguments using dynamic multi-pooling convolution neural network in pipelined fashion.
    
    \item JRNN ~\cite{nguyen-etal-2016-joint-event}: Exploit bidirectional RNN models and also consider event-event and event -argument dependencies in their model. 
    
    \item JMEE ~\cite{liu-etal-2018-jointly}: Use GCN model with highway network and self-attention for joint event and argument extraction.
    
    \item DBRNN ~\cite{Sha2018JointlyEE}: Add dependency arcs over bi-LSTM network to improve event-extraction.
    
    \item Joint3EE~\cite{Nguyen_Nguyen_2019}: Propose to share common encoding layers to enable the information sharing and decode trigger, argument and roles separately.
    
    \item GAIL~\cite{Zhang2019JointEA}: Propose an inverse reinforcement learning method using generative adversarial network (GAN).
    
    \item TANL~\cite{tanl}: Employ a sequence generation based method for event extraction.
    
    \item TEXT2EVENT ~\cite{lu-etal-2021-text2event}: Propose a sequence to structure network and infuse event schema by constrained decoding and curriculum learning.
    \item PLMEE ~\cite{Yang2019ExploringPL} Propose a method to automatically generate labelled data and try to overcome role overlap problem in EE task.
   
\end{enumerate}

\begin{table*}[ht]
\center\fontsize{12}{14}\selectfont
\setlength\tabcolsep{3pt}
\begin{tabular}{l|ccc|ccc|ccc|ccc}
\hline
\multirow{2}{*}{Model} & \multicolumn{3}{c|}{\begin{tabular}[c]{@{}c@{}}Trigger \\ Identify (TI)\end{tabular}} & \multicolumn{3}{c|}{\begin{tabular}[c]{@{}c@{}}Trigger \\ classify (TC)\end{tabular}} & \multicolumn{3}{c|}{\begin{tabular}[c]{@{}c@{}}Argument\\ Identify (AI)\end{tabular}} & \multicolumn{3}{c}{\begin{tabular}[c]{@{}c@{}}Argument-Role \\ Classify (ARC)\end{tabular}} \\ \cline{2-13} 
                        & P                         & R                        & F1                        & P                         & R                        & F1                        & P                         & R                         & F1                       & P                           & R                           & F1                         \\ \hline
JointBeam               & 76.9                      & 65.0                     & 70.4                      & 73.7                      & 62.3                     & 67.5                      & 69.8                      & 47.9                      & 56.8                     & 64.7                        & 44.4                        & 52.7                       \\ 
 DMCNN$^*$                   & 80.4                      & 67.7                     & 73.5                      & 75.6                      & 63.6                     & 69.1                      & 68.8                      & 51.9                      & 59.1                     & 62.2                        & 46.9                        & 53.5                       \\ 
JRNN                    & 68.5                      & 75.7                     & 73.5                      & 66.0                      & 73.0                     & 69.3                      & 61.4                      & 64.2                      & 62.8                     & 54.2                        & 56.7                        & 55.4                       \\ 
DBRNN                   & \multicolumn{3}{c|}{-}                                                          & 74.1                      & 69.8                     & 71.9                      & 71.3                      & 64.5                      & 67.7                     & 66.2                        & 52.8                        & 58.7                       \\ 
JMEE                    & 80.2                      & 72.1                     & 75.9                      & 76.3                      & 71.3                     & 73.7                      & 71.4                      & 65.6                      & 68.4                     & \textbf{66.8}                        & 54.9                        & \textbf{60.3}                       \\ 
Joint3EE      & 70.5             & 74.5            & 72.5            & 68.0                      & 71.8                     & 69.8                     & 59.9                     & 59.8                     & 59.9                    & 52.1                       & 52.1                       & 52.1                      \\ 
GAIL      & 76.8             & 71.2            & 73.9            & 74.8                      & 69.4                     & 72.0                     & 63.3                     & 48.1                     & 55.1                    & 61.6                       & 45.7                       & 52.4                      \\

PLMEE$^*$      & 84.8             & 83.7            & 84.2            & 81.0                      & 80.4.4                     & 80.7                     & 71.4                     & 60.1                     & 65.3                    & 62.3                       & 54.2                       & 58.0                      \\ 

TANL      & -             & -            & 72.9            & -                      & -                     & 68.4                     & -                     & -                     & 50.1                    & -                       & -                       & 47.6                      \\ 
\shortstack{TANL$_{multi}$}      & -             & -            & 71.8            & -                      & -                     & 68.5                     & -                     & -                     & 48.5                    & -                       & -                       & 48.5                      \\ 
TEXT2EVENT      & \multicolumn{3}{c|}{-}           & 69.6                      & 74.4                     & 71.9                     & \multicolumn{3}{c|}{-}                     & 52.5                       & 55.2                       & 53.8                      \\ \hline

\textbf{\shortstack{PESE$_{avg}$}}      & \textbf{95.3}             & \textbf{85.7}            & \textbf{90.2}            & \textbf{88.3}                      & \textbf{78.8}                     & \textbf{83.4}                     & \textbf{73.1}                     & 65.5                     & \textbf{68.9}                    & 61.9                       & 56.2                       & 58.4                      \\ 

\textbf{\shortstack{PESE$_{best}$}}      & \textbf{96.1}             & \textbf{86.1}            & \textbf{90.6}            & \textbf{89.4}                  & \textbf{79.5}                     & \textbf{84}                     & \textbf{74.1}                     & \textbf{66.6}                     & \textbf{69.8}                    & 63.3                       & \textbf{57.3}                       & 59.3                      \\ \hline
\end{tabular}
\caption{Performance comparison of our model against the previous state-of-the-art methods. \say{*} marked refers to the pipeline models and the remainings follow the joint learning approach }
\label{table:result}
\end{table*}

\section{Results \& Discussion}

Table~\ref{table:result} reports the overall performance of our proposed model(called PESE) compared to the other state-of-the-art EE models. We show the average scores over $4$ runs of the experiment in row PESE$_{avg}$. The row named PESE$_{best}$ describes our best $F_1$ scores in each subtask. We can see that, in TI, TC and AI task our model outperforms all the baseline models by a significant margin. Besides, for the argument-role classification (ARC) task our model achieves competitive results. The result table deduces some important observations: (1) In the TI task our model PESE$_{avg}$ outperforms all the baseline models and beat the second best model (\textbf{PLMEE}) by $6\%$ higher $F_1$ score. (2) Similarly, in the case of TC our model achieves the best performance by outperforming the second best model (\textbf{PLMEE}) by $2.7\%$ higher $F_1$ score. Moreover, the performance of our model in the trigger classification (TC) task is better than the best models that work specifically on TC subtask~\cite{xie-etal-2021-event}~\cite{tong-etal-2020-improving}. (3) However, the $F_1$ score of TC is reduced by more than $6\%$ compared to TI in both PESE$_{avg}$ and PESE$_{best}$ which indicates that in some cases, the model can correctly detect the trigger words but fails to identify the proper event types. In the ACE$2005$ dataset, among $33$ event types approximately $50\%$ events appear less than $100$ times. This imbalance in the training set may be a reason behind this fall in the $F_1$ score. (4) In the case of AI, our model achieves the best performance among all the baseline models achieving an average $F_1$ score of $68.9\%$. In the ACE2005 dataset, the maximum length of an argument is $38$ whereas the maximum length of a trigger is just $7$. It seems that the arguments with a long sequence of words and overlapping entities make the AI task more complex compared to the TI task where event triggers are mostly one or two words long. (5) In the ARC task, our proposed model achieves an average $F_1$ score of $58.4\%$ and is positioned third among all the reported baseline models. Our best result PESE$_{best}$ yields $F_1$ score of $59.3\%$ and only $1\%$ less than the best result (\textbf{JMEE}). However, without the infusion of any event-ontology information, we consider this end-to-end performance quite promising.
To further explore our model's effectiveness, we do some comparative experiments on the test dataset and report the performance on both single-event and multi-event scenarios in Table~\ref{table:n>=1}.

\begin{table}[ht]
\fontsize{10.5}{12}\selectfont
\begin{tabular}{|c|l|c|c|}
\hline
\multicolumn{1}{|l|}{\textbf{Item}}                                          & \textbf{Model} & \multicolumn{1}{l|}{\textbf{Count = 1}} & \multicolumn{1}{l|}{\textbf{Count \textgreater 1}} \\ \hline
\multirow{4}{*}{TC}                                                          & JMEE           & 75.2                                    & 72.7                                               \\ \cline{2-4} 
                                                                             & JRNN           & 75.6                                    & 64.8                                               \\ \cline{2-4} 
                                                                             & DMCNN          & 74.3                                    & 50.9                                               \\ \cline{2-4} 
                                                                             & \textbf{PESE}  & \textbf{82.6}                           & \textbf{84.1}                                      \\ \hline
\multirow{2}{*}{AI}                                                          & DBRNN          & 59.9                                    & 69.5                                               \\ \cline{2-4} 
                                                                             & \textbf{PESE}  & \textbf{65.3}                           & \textbf{71.4}                                      \\ \hline
\multirow{2}{*}{\begin{tabular}[c]{@{}c@{}}Argument \\ Overlap\end{tabular}} & BERD           & -                                    & 60.1                                               \\ \cline{2-4} 
                                                                             & \textbf{PESE}  & -                           & \textbf{74.3}                                      \\ \hline
\multirow{4}{*}{ARC}                                                         & JMEE           & \textbf{59.3}                           & 57.6                                               \\ \cline{2-4} 
                                                                             & DMCNN          & 54.6                                    & 48.7                                               \\ \cline{2-4} 
                                                                             & DBRNN          & 54.6                                    & 60.9                                               \\ \cline{2-4} 
                                                                             & \textbf{PESE}  & 54.1                                    & \textbf{61}                                        \\ \hline
\end{tabular}
\caption{Performance of our model with varied number of event records.}
\label{table:n>=1}
\end{table}

\subsection{Multiple Event Scenario:} Similar to previous works ~\cite{liu-etal-2018-jointly}~\cite{xie-etal-2021-event}, we divide the test sentences based on the number of event-triggers present and separately perform an evaluation on those sentences. In both single and multi-trigger scenarios, the model performs greater than $90\%$ in event type identification task. Interestingly, in the case of trigger classification (TC) also, the model performs comparatively better in multi-trigger instances, which presumes the effectiveness of our model in capturing the inter-event dependencies inside sentences. 

\subsection{Shared Argument Scenario} We also investigate our model's performance on the shared argument scenarios. In the ACE2005 dataset, an event instance may contain multiple arguments, or an argument phrase can be shared by multiple event instances. Compared to \textbf{DBRNN}, our model performs better in both single-argument and multi-argument scenarios.

\subsection{Overlapping Argument Phrases}
There are instances where parts of an entity phrase are considered as different arguments. For example, \textit{former Chinese president} is an \textit{Person} type argument whereas \textit{Chinese} is an \textit{GPE} type argument. When all the arguments inside a sentence are distinct, our model achieves $80.6\%$ $F_1$ score in argument phrase identification. Alternatively, in the presence of overlapping arguments, the $F_1$ score is  $74.3\%$, which is quite better than the results reported by BERD model~\cite{DBLP:conf/acl/Xi0ZWJW20}.

\subsection{Identifying Multiple Roles} Our model yields $F_1$ score of $54.1\%$ when each event mention has only one argument-role record within a sentence. In the presence of multiple argument-role information, the $F_1$ score is $61\%$. All the results are reported in Table~\ref{table:n>=1}  Similar to ~\cite{Yang2019ExploringPL}, we also consider the cases when one specific argument has single or multiple role information inside a sentence. For single role type, the model achieves $82.4\%$ $F_1$ score, and for multiple role instances, the corresponding $F_1$ score is $54.7\%$.

\subsection{Ablation Study}
To investigate the effects of external features employed in our model, we report the ablation study observations in Table~\ref{table:ablation}. We see that entity-type information is very critical for end-to-end event extraction. It improves the F1 score on each subtask very significantly. The quantitative scores also validate the use of pos-tag and dependency-tag features. The use of character-level features also gives us tiny improvements in the model performance. 

\begin{table}[ht]
\fontsize{10}{11.5}\selectfont
\begin{tabular}{|l|cccc|}
\hline
\multicolumn{1}{|c|}{\multirow{2}{*}{Model variation}} & \multicolumn{4}{c|}{F1-score}                                                     \\ \cline{2-5} 
\multicolumn{1}{|c|}{}                                 & \multicolumn{1}{c|}{TI} & \multicolumn{1}{c|}{TC} & \multicolumn{1}{c|}{AI} & ARC \\ \hline
PESE model                                        & \multicolumn{1}{c|}{\textbf{90.2}} & \multicolumn{1}{c|}{\textbf{83.4}} & \multicolumn{1}{c|}{\textbf{68.9}} & \textbf{58.4}  \\ \hline
- gold std. entity feat                                & \multicolumn{1}{c|}{84.7} & \multicolumn{1}{c|}{77.7} & \multicolumn{1}{c|}{62.6} & 51.5  \\ \hline

- pos tag feat                                         & \multicolumn{1}{c|}{86.9} & \multicolumn{1}{c|}{79.8} & \multicolumn{1}{c|}{66.1} & 55.2  \\ \hline
- dep feat                                             & \multicolumn{1}{c|}{87.4} & \multicolumn{1}{c|}{81.1} & \multicolumn{1}{c|}{66.9} & 55.7  \\ \hline
- char feat                                            & \multicolumn{1}{c|}{89.7} & \multicolumn{1}{c|}{81.9} & \multicolumn{1}{c|}{67.1} & 56.3  \\ \hline

- all external feat                                            & \multicolumn{1}{c|}{82.3} & \multicolumn{1}{c|}{75.9} & \multicolumn{1}{c|}{61.3} & 49.9  \\ \hline


\end{tabular}
\caption{Ablation of external features on model performance.}
\label{table:ablation}
\end{table}

\section{Related Works}
Based on the ACE2005 guidelines the task of EE is the composition of three to four subtasks corresponding to different aspects of the event definition~\cite{Nguyen_Nguyen_2019}. A large number of prior works on EE only focus on some specific subtasks like: event detection~\cite{Nguyen2015EventDA} ~\cite{xie-etal-2021-event}~\cite{tong-etal-2020-improving} or argument extraction~\cite{Wang2019HMEAEHM}~\cite{Zhang2020ATA}~\cite{Ma2020ResourceEnhancedNM}. The models that are capable of extracting the complete event structure are categorized in mainly two ways: (1) pipelined-approach~\cite{ahn-2006-stages}~\cite{ji-grishman-2008-refining}~\cite{Hong2011UsingCI}~\cite{Huang2012ModelingTC}~\cite{Chen2015EventEV}~\cite{Yang2019ExploringPL} and (2) joint modeling approach~\cite{McClosky2011EventEA}~\cite{li-etal-2013-joint}~\cite{yang-mitchell-2016-joint}~\cite{liu-etal-2018-jointly}~\cite{Zhang2019ExtractingEA}~\cite{Zheng2019Doc2EDAGAE}~\cite{Nguyen2019EndtoendNR}. Recently, methods like question-answering~\cite{Du2020EventEB}~\cite{Li2020EventEA}, machine reading comprehension~\cite{Liu2020EventEA}, zero shot learning~\cite{Huang2018ZeroShotTL} are also used to solve the EE problem. Some of the recent works that follow sequence generation approach for event extraction also achieve promising results~\cite{tanl}~\cite{Du2021GRITGR}. Among the previous methods the closest to our approach is TEXT2EVENT~\cite{lu-etal-2021-text2event} that also generates the event structure from sentences in end-to-end manner. But they generates the event representations in token by token format that means in each time step the model generates one single token. Whereas our model generates one single event frame per time step which is more realistic in end-to-end event structure extraction.

\section{Conclusion}
In this paper, we present a joint event extraction model that captures the event frames from text, exploiting intra-event and inter-event interactions in an end-to-end manner. Unlike other methods that consider EE as a token classification problem or sequence labeling problem, we propose a sequence-to-tuple generation model that extracts an event-tuple containing trigger, argument, and role information in each time step. The experimental results indicate the effectiveness of our proposed approach. In the future, we plan to use cross-sentence context in our model and infuse event ontology information to improve our performance.

\bibliography{acl_latex}
\bibliographystyle{acl_natbib}




\end{document}